\documentclass[conference]{IEEEtran}
\IEEEoverridecommandlockouts

\usepackage{graphicx}
\usepackage{amsmath,amsfonts,amssymb}
\usepackage{algorithm}
\usepackage{algorithmic}
\usepackage{url}        
\usepackage{amsmath}    
\usepackage{cite}
\usepackage{textcomp}
\usepackage{xcolor}

\begin{document}
\title{Blockchain Solutions for Multi-Agent Robotic Systems: Related Work and Open Questions}

\author{
\IEEEauthorblockN{Ilya Afanasyev\textsuperscript{1}, Alexander Kolotov\textsuperscript{1}, Ruslan Rezin\textsuperscript{1}, Konstantin Danilov\textsuperscript{1}, Alexey Kashevnik\textsuperscript{2}, Vladimir Jotsov\textsuperscript{3}}
\IEEEauthorblockA{\textsuperscript{1}Innopolis University, Innopolis, Russia \\
\textsuperscript{2}ITMO University, St.Petersburg, Russia \\
\textsuperscript{3}University of Library Studies and Information, Sofia, Bulgaria \\
\{i.afanasyev, a.kolotov, r.rezin, k.danilov\}@innopolis.ru, alexey@iias.spb.su, v.jotsov@unibit.bg\\}
}

\maketitle

\begin{abstract}
The possibilities of decentralization and immutability make blockchain probably one of the most breakthrough and promising technological innovations in recent years. This paper presents an overview, analysis, and classification of possible blockchain solutions for practical tasks facing multi-agent robotic systems. The paper discusses blockchain-based applications that demonstrate how distributed ledger can be used to extend the existing number of research platforms and libraries for multi-agent robotic systems.
\end{abstract}

\section{Introduction}
\label{introduction}

The perspectives of decentralization and immutable records make blockchain probably one of the most powerful technological innovation in recent years.
To be ubiquitous the blockchain needs to overcome the following issues: scalability, latency and low throughput \cite{edChain}.

Permissionless public blockchains have nodes (i.e. computers in the network) that maintain a shared version of the records by storing on each node a complete information on the blockchain. This ensures the blockchain permanence from cyber attacks, which cannot destroy it just by taking out a central server. Most publicly available blockchains also have very reliable means of ensuring data integrity. The blockchain can only be updated by adding a new block without deleting and modifying the existing blocks. Therefore, it is necessary to maintain the integrity of the order of transactions. The blockchain does this through a consensus mechanism. The most widely used consensus mechanism applies proof of work (POW) algorithm, where a miner must solve a complex cryptographical operation with a large number of blocks at high-speed, in a competitive environment. Moreover, the POW algorithm requires that the majority of the participating nodes approve the transaction. In a decentralized network, the hijack of the computing power of the majority of the node looks economically non-viable \cite{edChain}. However, since the blockchain grows with more users and a higher number of transactions, the transaction validation makes the transaction process slower. 

Let’s summarize the key points of blockchain technology addressing them to operation of the multi-agent systems (further - robotics agents): 
\begin{itemize}
\item The blockchain is an append only database. As soon as data are included in the database it cannot be changed.
\item The database shapes the state of the blockchain. Examples of the blockchain state: records of all accounts balances, snapshot of memory controlled by executable agents. 
\item The database is distributed among nodes. Every node keeps full copy of the database. 
%
%
%
\item The node is an agent in the blockchain network.
\item The node could generate records to change the state of the blockchain. 
\item The node is responsible for transferring all incoming data arrived from another node to all its neighbors.
\item All nodes are connected through peer-to-peer communication channels. 
\item Some nodes must play the validator role.
\item The validators verify correctness of records to change blockchain state and approve them (e.g. by combining the records in blocks, linking the blocks together and sending new blocks to the neighbors). 
\item Only validated records are applied on all the nodes to build the current state of the blockchain.  
\end{itemize}

Concluding the information above, let’s highlight the strengths of the blockchain technology for multi-agent systems: 
\begin{itemize}
\item Data availability is achieved through multiply duplication of data and communication.
\item Consistency of data is achieved through data validation and strict rules of changes appliance. 
\item No way to remove or change the data stored in the blockchain.
\item Economic or reputational incentive forces nodes to not violate the validation rules. 
\end{itemize}

The relevance of the study is explained by the priority in the development of intelligent multi-agent robotic systems capable of performing various tasks with a high degree of autonomy. Information support for the interaction of robot groups has a particular importance for operations in uncertainty conditions, external disturbances and environmental changes. Successful solution of the group interaction problem together with recording the interaction history and performing the verification task by the distributed ledger technology (blockchain), can increase the efficiency of interaction between groups of robots and expand the possibilities of their applications.

The rest paper is structured as follows. The Section \ref{related_papers} introduces the present state of scientific and engineering development in the blockchain-based multi-agent systems. Section \ref{state-of-the-art} describes and classifies the most typical cases, which we identified for blockchain-based robotics applications. Finally, we open the dialog for discussion in the Section \ref{discussion}. 

\section{Related paper analysis}
\label{related_papers}

In recent years, research and development in the field of multi-agent robotic systems have become increasingly popular. Considerable attention is paid to the development of distributed planning systems for robot coalitions and algorithms for distributed control of robotic networks \cite{bullo2009, liu2015, peng2015, savkin2016, wang2016}. The section presents a related work analysis in the area of blockchain utilization for multi-agent robotic systems. 

The paper \cite{Pawlak2018} considers the utilization of intelligent agents and multi-agent system concept for Auditable Blockchain Voting System. The aim of such system is integration of e-voting processes with blockchain technology. The main problem of the electronic voting is the low trust level of the respondents to the system since the results can be simply changed. Authors propose to utilize the multi-agent approach to increase the trust level in e-voting system. 
In scope of the paper \cite{Skowronski2019} author propose a new family of cyber-physical systems. The proposed coupling of the data storage mechanisms, communication and consensus protocols, allows for deployment of self-sustaining cyber-physical environments in which all the mission-critical aspects in both the cyber and physical layer are effectively incentivized, coordinated, and maintained. The following benefits the blockchain technology for cyber-physical systems utlization are highlighted by authors: decentralized, replicated data storage that improves rigidness of the system; ability for anyone to join; participants reach consensus within a trustless environment; ability to maintain trust among initially unknown agents; distributed execution of atomic applications; state proposals are handled exactly as the majority of the network commands; transparency and immutability; agents remain fully autonomous, they are in complete control over their identities and private keys; blockchain data is complete, consistent in the long run and available.

Authors of the papers \cite{casino2018, Wang2019} describe in details the application of the blockchain technology for Internet of Things (IoT). Authors mentioned that IoT concept transforms human life and unleash enormous economic benefits. But however, inadequate data security and trust level are seriously limiting its wide application. Authors propose a survey where they describe in detail the application of the blockchain technology to Internet of Things environments. Similarly, IoT and Smart and Software-defined Buildings (SSDB) technologies and their cooperation in implementing Smart Spaces are considered in the article \cite{mazzara2019}, which offers a reference architecture for IoT infrastructure and assumes that the blockchain may be the highway for its future development. Developing the idea of a system of smart buildings, the authors in the paper \cite{lazaroiu2017} integrate subsystems, such as intelligent networks, services, buildings and household appliances, into models of smart districts and even smart cities, where these subsystems efficiently interact, connect and control remotely by using blockchain technology to achieve a better quality of life, sustainability, energy conservation and socio-economic system development.

Since the development of robotics, IoT, big data processing, automation and distributed register technology (DLT) leads to the fourth industrial revolution, the interaction of the “smart factory” components within the company and with other Industrial IoT participants providing trust, control over the distribution of resources and products, is investigated in the works \cite{teslya2017, kapitonov2018}.
The paper \cite{casino2018, Casado-Vara2018} discusses the utilization of the blockchain technology to supply chain. Generally supply chain is the multi-agent system where every supplier has own behaviour model and goals. In scope of the paper \cite{Casado-Vara2018} authors propose new model of supply chain based on blockchain concept. This new model enables the concept of circular economy and eliminates many of the disadvantages of the current supply chain.

The recent review of the state of the art in the application of blockchain technology to multi-agent systems is presented in the paper \cite{calvaresi2018}, which describes some cases of multi-agent systems, although it did not classify the multi-agent systems by purpose and did not present multi-robot systems using the blockchain \cite{kapitonov2017, danilov2018, teslya2018, Kashevnik2018}.
~
The research \cite{kapitonov2017} focuses on the organization of a blockchain-based protocol for multi-agent coordination and control in the context of Unmanned Aerial Vehicles (UAVs). 
~
The article \cite{danilov2018} considers a blockchain consensus protocol, which additionally includes a validation procedure of liability execution to prevent payment transactions to questionable service providers. The proposed methodology of a liability execution for agent-based service providers in a decentralized trading market uses a Model Checking method. As the proof-of-concept, the methodology was implemented in an application that simulates the work of a taxi with the following liability validation at the end of a completed scenario.
~
The methodology for building the cyber-physical smart space proposed in \cite{teslya2018} describes how to form and operate coalitions of intelligent robots using knowledge processors and information stored on the blockchain. To ensure the interaction of heterogeneous robots in the cyber-physical space, an ontology can be applied, which represents the knowledge and competencies of robots in the system. The methodology provides the fast information exchange between coalition members and smart contracts for the distribution of sensory, computational, control and service tasks between intelligent robots, embedded devices and information resources.
~
Similarly, in the paper \cite{Kashevnik2018} the cyber-physical-social system is considered that is operated based on a smart space technology and blockchain concept. Authors propose case studies that include mobile robots and human participation. Interaction between robots and humans is implemented based on ontology-based publication/subscription mechanism and all information exchange traffic has been monitored and key information is stored in the blockchain network. 

Another solution that integrates robotics and blockchains is studied in the paper \cite{lopes2019}, which offers a modular architecture that utilizes the RobotChain framework \cite{ferrer2018b} as a decentralized ledger for registering robotic events, smart contract technology for managing robots, and Oracle for processing data of any types.
The modular architecture can be used in various contexts, such as production, network management, robot management, etc., since it is easy to integrate, adapt, maintain and expand for new domains. 
The examples of applications include: (a) task distribution between a robot network; (b) supporting methodology to assist robots in their task performance if they have cannot execute or need the specific information (e.g. a robot could not know what are the objects whereas other robots could); (c) robot productivity estimation or workability issue detection that can be valuable for industrial applications; (d) voting consensus for swarm robotics. What is more, tokens can be dropped to speed up the validation process or replaced by a reputation system for task management and consensus, since monetary value is no longer meaningful for private blockchain-based networks.

The article \cite{shukla2018} discusses the implementation of intelligent cyber-physical systems as multi-agent systems with the ability to schedule tasks by agents. In such multi-agent systems, the plan execution protocol should lead to proper completion and streamlining of actions, despite their distributed execution. However, in unreliable scenarios, there is a probability that agents will not follow the protocol due to faults or due to malicious reasons that lead to plan failure. In order to prevent such situations, the authors \cite{shukla2018} proposed the proper plan execution by agents through smart contracts that will help accomplish the task even in untrusted environment. The researchers \cite{shukla2018} developed the architectures in which smart contracts can be automatically generated from derived plans, as a result the entire system of smart, intelligent agents can be fully automated and the overall architecture will seamlessly integrate agents into one cyber-physical system.

\section{The classification of the blockchain-based robotics applications}
\label{state-of-the-art}

\begin{figure}[!htbp]
    \centering
    \includegraphics[width=\columnwidth]{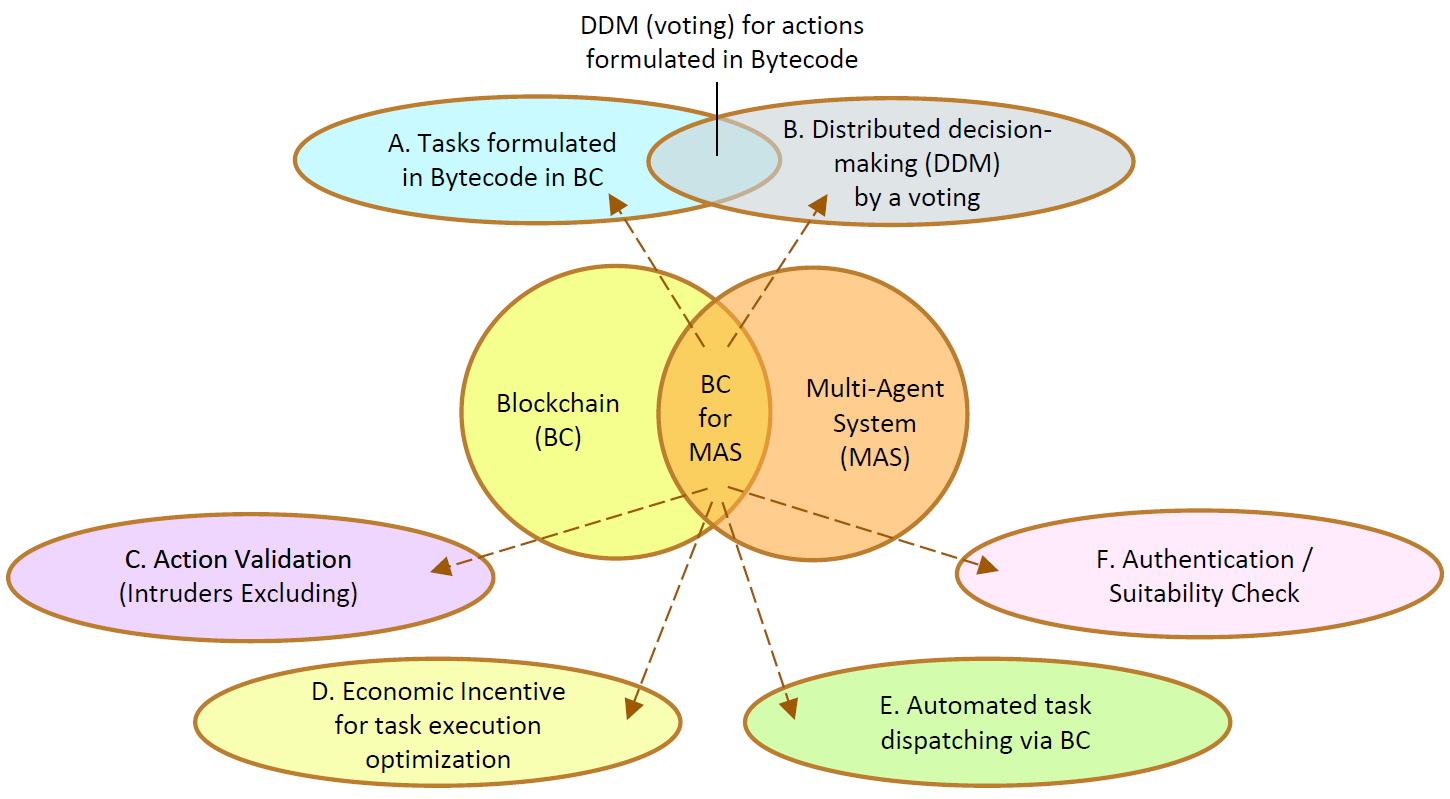}
    \caption{The classification of the most typical cases for using blockchain (BC) in multi-agent systems (MAS) concerning robotics applications}
    \label{fig01}
\end{figure}

This section classifies the most typical cases for blockchain-based multi-agent systems concerning robotics applications, which we discovered during our study. The identified cases are shown in Fig. \ref{fig01}.

\subsection{Logging agent actions with a bytecode distribution}

Consider an environment which consists of tons of very general agents. The agent has no predefined control program
but configured as so could execute a batch of commands provided in form of a bytecode or similar. For example, an 
agent is a differential drive robot with sensors and gripper so the sequence of commands to execute could be: 
follow the wall until discover the room with the green floor, find the yellow box and bring it to the target point.
~
The word bytecode is used here as a description of the platform-independent executable code. In other words, the same byte code will lead to the same sequence of command execution on any agent. Thus, at the moment of sending the bytecode, it is not necessary to know the exact hardware platform of the agent that must execute the commands specified in the bytecode.

The blockchain could be used here to distribute such bytecodes to the agents:

- The system which generates the tasks should not be connected to the agent directly. The peer-to-peer network
%
%
      is used to deliver the message.

- A message will be delivered even if an agent is turned off.

- The agent is able to inform constantly about its state changes (e.g. "moved forward for 1 meter", "picked the box up"). The state is stored in the blockchain therefore it could be recovered quickly in case of the agent recovery after halt.

- The delayed bytecode execution could be scheduled.

- Two or more command consequences could be automatically queued to be executed by the agent.

\subsection{Distributed decision making by a time-limited voting}
Distributed decision making is a very important task in multi-agent systems, especially in SWARM systems. Blockchain application for this task had been already proposed \cite{ferrer2018}. However, proposed solution was based on the idea of sending coins to some "proposal" addresses. In general, blockchain provides advanced technologies, which can solve this task in a better way, specifically, through smart-contracts, for instance in Ethereum.

One can develop smart-contract(s) to build infrastructure, providing ability to implement polls with complicated behavior, like time-limited voting or vote delegating.

By combining this approach with the previous case, we can get another interesting solution. The usage of smart-contract agents can propose some actions, formulated in bytecode. Other agents can vote for actions. Finally, collaboratively generated scenario can be gathered from the smart-contract.

\subsection{Action validation to exclude intruders or faulty agents}

Agents could be used to validate actions, location or poses of each other. Let’s consider the swarm where agents perform some movements as per the general goal. Periodically the agents send information about the actual sensor information and the agent location based on odometry. Sometimes the agent could start misoperation and send incorrect data. Information received from other peers could be used to get to consensus that the agent works improperly and recovery procedure could start. In this case co-evolution scenarios could be further applied. The consensus based on the information received from other agents could be also used aiming to identify the robot that misbehaves intentionally e.g. it was hacked or injected by an intruder. 
~
In order to resolve the issue with performance of validators the approach which is called \textit{Sharding} could be used \cite{edChain}. The data from the agents is combined based on their location: thus a shard is built. Validators are negotiated for the shard that is why the volume of the information to treat is reduced significantly.

\subsection{Economic Incentive for task execution optimization}

The financial part of the blockchain can be used as a basis for the approach to Multi-robot exploration controlled by a market economy proposed by researchers \cite{zlot2002} from Carnegie Mellon University, where multirobot mapping was performed using the market approach to robot team coordination and to maximize the information acquisition at minimal cost. This multi-robot exploration system demonstrated reliability and adaptability to a dynamic environment and to the loss of colony members in addition to its ability to withstand communication losses and failures. The researchers found that by allowing robots to negotiate using the market architecture. The environment exploration effectiveness was improved several times for the robot team. Although the algorithm was designed to minimize the distance traveled during the exploration, the usage of time-based costs instead of distance-based costs would lead to much faster exploration. Such an approach can simplify setting priorities for some types of compared tasks to others within the market, for example, if there are other mission objectives in addition to the exploration \cite{zlot2002}.

\subsection{Automated task dispatching via blockchain}

The distributed consensus existing as a part of the blockchain-based solutions could be used to perform  task dispatching among competing agents for the assignment to complete the task.

The dispatching code is written in form of the smart contract stored in the blockchain:

\begin{enumerate}
\item The customer sends a request to perform a task to the dispatcher of the smart contract.
\item The dispatcher notifies the agents about the new request.
\item The agents send agreement to perform the task to the blockchain through peer-to-peer network. 
\item Blockchain validators define the order of the agreements as per the fee the particular agent is paying for handling of the agreement.
\item The first agreement received by the dispatcher is confirmed by the code of the smart contract and the details of the order are provided to the corresponding agent.
\end{enumerate}

Thus, the market will regulate the selection of agents to be dispatched to complete the task. The agents, which were provided with a cost optimization strategy, could pay more to be selected by validators. This should result in appearing the most efficient and stable service providers. 

The case study towards the creation of such a multi-agent system is presented in the study \cite{danilov2018}], in which an automated dispatching taxi scenario is implemented with the liability execution validation (by comparing the traveled route with a map by a validator). 

\subsection{Authentication / suitability check}

There are situations when agents who do not trust each other but use a shared physical resource, or when charging and billing process may put users at risk of hacker’s attack or compromise the privacy and/or reveal their confidential information and/or location. 

One of the examples is a replaceable battery for electric vehicles (EV). Vehicles use maintenance stations to replace batteries. Both the maintenance and the vehicle’s owner are interested to ensure that the battery meets the quality issues and that the existing level of battery amortization is declared, since fraud with this shared resource can lead to a loss of profit and trust in such a service. For EV owners, the battery in- formation and transaction correctness, openness, traceability and immutability is difficult to get guarantee in traditional centralized system. The trust lacking between EV owners and the maintenance station can cause a big challenge to the EV market development. The blockchain-based solutions can be used to provide battery authentication services. The smart contract code stored on the blockchain cannot be changed, hence battery amortization state is available to any deal participants. The maintenance station and the vehicle can connect to any node in the blockchain aiming to perform the battery check, so "the man-in-the-middle attack" becomes difficult to perform the data replacement. 

The case study of such a system was proposed in the paper \cite{hua2018}, where the issue of battery swapping and trust lacking was resolved through application of a decentralized blockchain system. Within this solution, both battery life-cycle information and all operation histories were permanently saved in the blockchain network. All key logics were driven by smart contracts, the battery price calculation and the digital currency exchange between EV owners and the station were realized by smart contracts automatically aiming to resolve the lack of trust issues.
~
Similarly, the proposed research \cite{dragos2019} discusses the new autonomous charging architecture without involving any human. It also provides a billing framework for Electric Autonomous Vehicles (EAVs) which allow cost-effective Machine-to-Machine (M2M) transactions using Distributed Ledger Technology (DLT), possessing greater resistance to hacker attacks and retaining confidential user information.

\section{Discussion}
\label{discussion}

At present, the approach to organizing an immutable distributed database that stores all relevant information and provides access to agents of a multi-agent robotic system, expanding the capabilities of the system as a whole, has the considerable interest within the Fourth Industrial Revolution concept. The key component of the approach is a distributed ledger technology (blockchain) associated with a virtual computer, which allows agents to interact through responsible “smart contracts” \cite{teslya2017, kapitonov2018}. On the one hand, the reliability of the agent is mainly determined by the reputational model. However, this allows you to determine the trust level to the agent only after the fulfillment of the agreed obligations. On the other hand, the automation of obligation fulfillment by an agent can provide a verification procedure that will allow to verify the liability execution (e.g., using formal methods \cite{danilov2018}). In addition, one of the goals of the implementation of the blockchain for a multi-agent system may be the increase of the interaction efficiency between agents by organizing more trusted information support.

The main idea of this paper is  to provide a guidance on how frameworks back-uped by blockchain solutions could be used to address practical tasks faced by multi-agent systems, especially for groups of mobile robots. The relevant reviewed studies show up that the blockchain could play a significant role in multi-agent system applications. The analysis of recent publications allow us to identify groups of tasks for blockchain-based multi-agent robotic systems, which we proposed for classification. Blockchain technologies could be used to extend existing number of platforms and libraries used by researchers or motivate them to use a common solution that is widespread and verified instead of putting efforts in developing their own code solutions to cover similar scenarios.

Based on the state-of-the-art analysis authors conclude that at the moment the promising task in the area of multi-agent systems is the development of methodologies, models and methods focused on an intelligent support of blockchain-based agent interactions that significantly increase the level of trust in such systems. The following tasks remain "opened" at the moment and need to be solved: 

\begin{itemize}
 \item Development of a conceptual model of information support for a group of robots during the task performance; 
 \item Development of a typical ontological model of a robotic system; 
 \item Development of a consensus protocol for a group interaction verification before launching a task based on the information from a distributed ledger;
 \item Development of a validation method for task performance by the robotic system;
 \item Development of multi-agent system architecture;
 \item Prototyping of a framework for an intelligent group of robots performing a collaborative task (as a proof-of-concept).  
\end{itemize}


\end{document}